\documentclass[letterpaper, 10 pt, conference]{ieeeconf}  

\IEEEoverridecommandlockouts                              

\makeatletter \let\NAT@parse\undefined \makeatother

\usepackage{graphics} 
\usepackage{epsfig} 
\usepackage{mathptmx} 
\usepackage{times} 
\usepackage{amsmath} 
\usepackage{amssymb}  

\usepackage{microtype} 
\usepackage{xcolor} 
\usepackage{censor}

\usepackage{graphicx}
\usepackage{booktabs}
\usepackage{subcaption}
\DeclareCaptionFont{eightpt}{\fontsize{8}{9.6}\selectfont}
\usepackage{bm}
\usepackage{mathtools}

\usepackage{algorithm}
\usepackage{algorithmic}

\usepackage[textsize=tiny, colorinlistoftodos]{todonotes}

\usepackage[
        pagebackref=false, 
        breaklinks=true, 
        colorlinks, 
        citecolor=black, 
        linkcolor=black,
]{hyperref} 
\usepackage[capitalise]{cleveref} 
\usepackage{fancyhdr}

\title{\LARGE \bf
Sampling-Based MuJoCo MPC for Double-Pendulum Sway Suppression on a Shipboard Crane
}

\author{Oscar Pang$^{\dagger}$, Lisa Coiffard$^{\dagger}$, Paul Templier$^{\dagger}$, Luke Beddow$^{\dagger}$, Kamil Dreczkowski$^{\dagger}$, Antoine Cully$^{\dagger}$
 \thanks{*This work is funded in part by the Defense Advanced Research Projects Agency (DARPA) Learning Introspective Control (LINC) program.}
 \thanks{$^{\dagger}$The authors are with the Adaptive and Intelligent Robotics Lab, Department of Computing, Imperial College London, Exhibition Rd, London, SW7 2BX, United Kingdom 
}%
 \thanks{Corresponding authors: Oscar Pang (\texttt{k.pang@imperial.ac.uk}) and Antoine Cully (\texttt{a.cully@imperial.ac.uk})}
}

\begin{document}

\maketitle

\thispagestyle{fancy}
\fancyhead{}
\fancyfoot{}
\fancyfoot[C]{%
  \rule{0.95\columnwidth}{0.4pt}\\[2pt]
  \footnotesize\textit{This work has been submitted to the IEEE for possible publication. Copyright may be transferred without notice, after which this version may no longer be accessible.}
}
\renewcommand{\headrulewidth}{0pt}

\begin{abstract}
Transferring heavy payloads in shipboard settings relies on efficient crane operation, limited by hazardous double-pendulum payload sway that wind and ocean-wave perturbations amplify.
Suppressing these oscillations on an underactuated crane is challenging for human operators, and existing control methods struggle in such settings: analytical model predictive control (MPC) relies on linearized or reduced-order models, while reinforcement learning (RL) requires extensive offline training and cannot adapt once deployed. 
Deploying a predictive controller on compute-constrained hardware, without extensive offline training or analytical model derivation, remains difficult.
We present a complete real-time control pipeline centered on the MuJoCo MPC framework, using a cross-entropy method planner to evaluate candidate action sequences directly through simulation rollouts. 
This sampling-based approach reconciles the competing objectives of dynamic target tracking and payload sway damping without an analytical model. 
We show that the controller runs effectively on embedded hardware and achieves significantly lower position error than PID and RL across all conditions, while remaining competitive on sway damping under base perturbation, and that it degrades gracefully under unmodeled physical discrepancies such as an added second payload.
\end{abstract}

\section{Introduction}

Shipboard crane operations face a persistent challenge: the coupled effects of crane arm 
acceleration and continuous base perturbations from ocean waves often cause payload sway during transfer. 
This sway often manifests as double-pendulum motion, where the payload oscillates both from the boom tip and about the hook-cable connection~\cite{vaughan_control_2010}.
The underactuated nature of the crane makes manual stabilization difficult, degrading payload positioning accuracy and posing risks to surrounding equipment, particularly when ship-motion-induced disturbances amplify the oscillations~\cite{ramli_control_2017, cao_review_2020}.
Suppressing these oscillations in real-time\footnote{We use \textit{real-time} in the operational sense of \cite{howell_predictive_2022}: the planner produces updated policies within a single control cycle, enabling continuous closed-loop operation without offline computation.} requires a controller that can anticipate disturbances while managing the competing demands of fast payload transfer and sway minimization.

\begin{figure}[tb]
\centering
\includegraphics[width=0.9\linewidth]{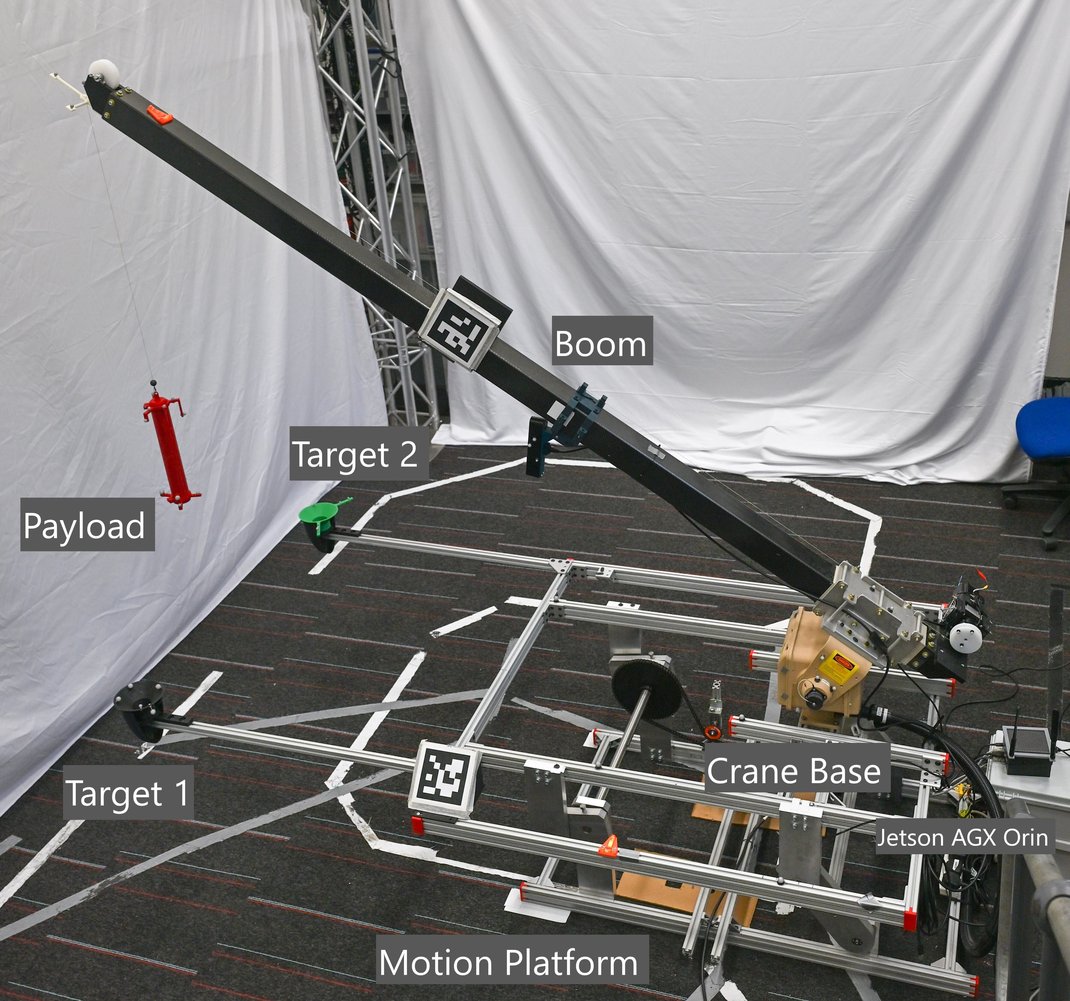}
\caption{Experimental setup of a boom-type crane mounted on a motion platform.}
\label{fig:motion_platform}
\end{figure}

\begin{figure*}[tb] 
\centering
\includegraphics[width=1.0\linewidth]{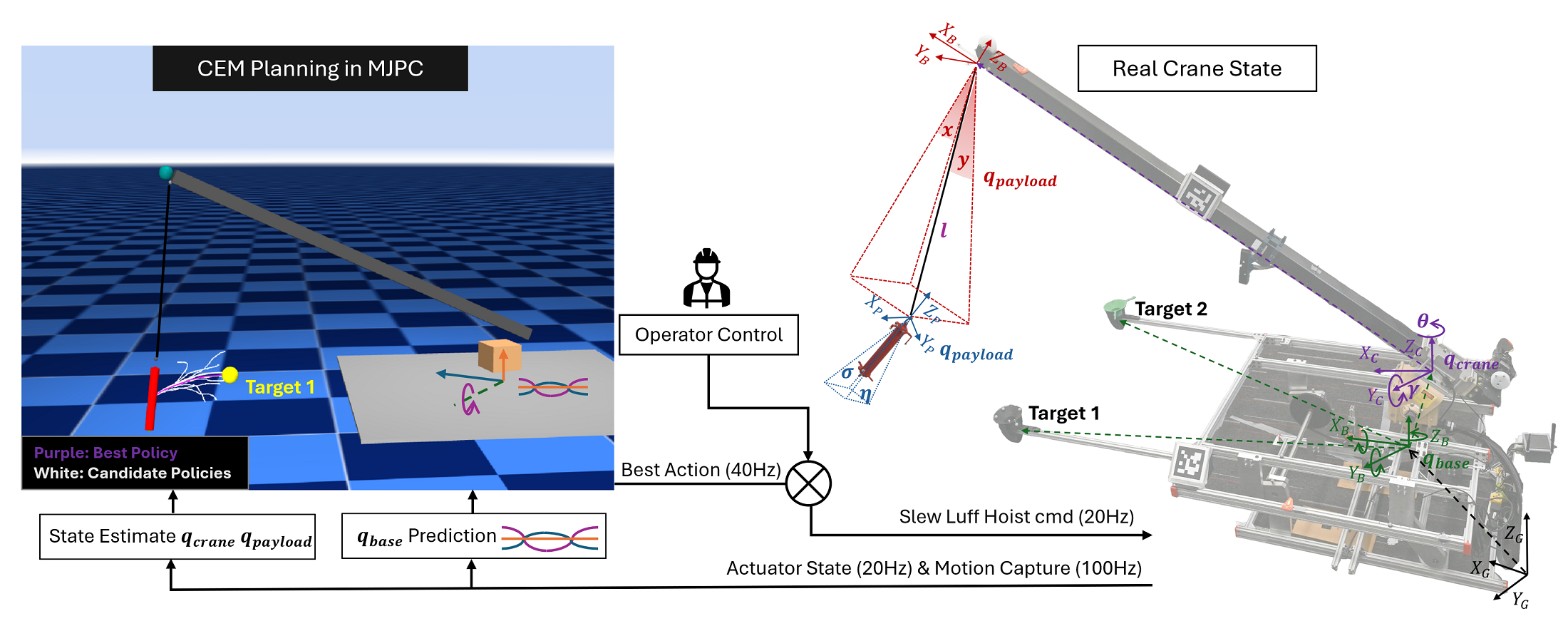}
\caption{MJPC crane controller architecture. \textit{Right:} the physical crane with its frames of reference (global, base, crane, boom tip, payload) and joint positions ($q_{\text{base}}$, $q_{\text{crane}}$, $q_{\text{payload}}$). \textit{Left:} at each step, the state estimates of $q_{\text{crane}}$ and $q_{\text{payload}}$ and the predicted $q_{\text{base}}$ sequence are fed to the CEM planner in MJPC, which samples candidate policies (white), rolls them forward through simulation, and selects the best (purple). The immediate action of the best policy is issued as the crane control command; an operator can override the planner via joystick ($\otimes$).}
\label{fig:system_architecture}
\end{figure*}

The field of crane anti-sway control is mature and well-studied~\cite{ramli_control_2017}, with approaches spanning input shaping~\cite{vaughan_control_2010}, linear and non-linear feedback~\cite{rigatos_nonlinear_2024, sun_nonlinear_2018} and model predictive control (MPC)~\cite{lin_constrained_2024, cao_nonlinear_2024, richter_realtime_2014}. 
These share a common dependency on analytically derived equations of motion, typically linearized or reduced in degrees-of-freedom (DoF) to remain computationally tractable~\cite{ramli_control_2017, cao_review_2020, das_safe_2025}.
For shipboard cranes under persistent base perturbations, these simplifications discard the coupled slew–luff dynamics, double-pendulum modes, or non-linear terms that dominate the system response in rough seas~\cite{cao_review_2020}.
Reinforcement learning (RL) sidesteps the modeling burden, but requires extensive offline training and, once deployed, cannot adapt its policy when the system dynamics change~\cite{akki_benchmarking_2025}.

Recent work in robotics treats general purpose physics engine as planning models for real-time control: candidate action sequences are rolled forward through simulation and the best-performing sequence is executed on hardware. MuJoCo MPC (MJPC)~\cite{howell_predictive_2022} provides an open-source framework for this paradigm, which has produced impressive results across robotic tasks, but its adoption has been confined to systems where high-fidelity models with calibrated actuator dynamics are readily available in physics engines~\cite{menagerie2022github, shaw2023leaphandlowcostefficient}.

We extend this paradigm to design a controller capable of payload stabilization in shipboard crane, where no such simulation model exists for the proprietary industrial system, and where persistent base perturbation couples directly into the unactuated payload dynamics. Our contributions are:

\begin{enumerate}
    \item We present an alternative approach to analytical-model MPC by planning directly over the full double-pendulum dynamics via physics-engine rollouts and sampling-based optimization. This sidesteps the need for linearization, DoF reduction, or the planner–stabilizer decomposition that existing crane control methods require.
    \item We validate our approach on a physical crane across multiple emulated sea-state conditions, benchmarking against PID and RL baselines. We show that it reconciles target tracking and active sway damping, achieving significantly lower position error than both baselines while remaining competitive on sway.
    \item We analyze the trade-off between CEM iteration budget and planning frequency for real-time deployment on compute-constrained hardware.

\end{enumerate}

\section{Background}
\label{sec:background}

\subsection{Problem Formulation}
\label{subsec:problem_formulation}

We consider a boom-type crane with three actuated joints: slew, luff, and hoist mounted on a motion platform that emulates shipboard motion at sea, which we assume to be periodic and of fixed amplitude, by generating disturbances to the crane base (\cref{fig:system_architecture}). A rigid cylindrical payload of mass $m$ is suspended from the boom tip and exhibits double-pendulum dynamics, swinging freely about the boom tip and the hook of the cable. 

The system state 
$\mathbf{x}_t = [\mathbf{q}^\top, \dot{\mathbf{q}}^\top]^\top \in \mathbb{R}^{26}$ comprises 13 generalized joint positions and their velocities:
\begin{equation}
\label{eq:system_state}
    \mathbf{q} = [\mathbf{q}_{\text{base}}^\top,\, 
    \mathbf{q}_{\text{crane}}^\top,\, 
    \mathbf{q}_{\text{payload}}^\top]^\top \in \mathbb{R}^{13},
\end{equation}
where $\mathbf{q}_{\text{base}} \in \mathbb{R}^{6}$ is the platform pose, $\mathbf{q}_{\text{crane}} \in \mathbb{R}^{3}$ the actuated crane joints, and $\mathbf{q}_{\text{payload}} \in \mathbb{R}^{4}$ the double-pendulum swing angles at the boom tip and hook. Only $\mathbf{q}_{\text{crane}}$ is directly actuated: the base motion acts as a measured disturbance and the payload states are unactuated.

\subsection{Sampling-Based Model Predictive Control}
\label{subsec:background-mpc}

We formulate the task of shipboard crane payload stabilization, which consists of target tracking and payload sway damping, as an optimal control problem over a receding horizon $H$. The objective is to find a sequence of control actions $\mathbf{u}_{t:t+H-1}^* = \{\mathbf{u}_t, \dots, \mathbf{u}_{t+H-1}\}$ that minimizes a cumulative cost:

\begin{equation}
\begin{aligned}
    \mathbf{u}_{t:t+H-1}^* = \arg\min_{\mathbf{u}_{t:t+H-1}} \quad 
    & \sum_{k=0}^{H-1} c(\mathbf{x}_{t+k}, \mathbf{u}_{t+k}) 
    + c_f(\mathbf{x}_{t+H}) \\
    \textrm{subject to} \quad 
    & \mathbf{x}_{t+k+1} = f(\mathbf{x}_{t+k}, \mathbf{u}_{t+k}), \\
    & \mathbf{u}_{\min} \leq \mathbf{u}_{t+k} \leq \mathbf{u}_{\max}, \\
    & \mathbf{x}_{t} = \hat{\mathbf{x}}(t).
\end{aligned}
\label{eq:mpc_objective}
\end{equation}

Following prior applications of simulation-based planning for control of physical hardware~\cite{li_drop_2025}, we adopt the cross-entropy method (CEM) \cite{rubinstein_cross-entropy_2004} on \eqref{eq:mpc_objective}.
Rather than optimizing the control sequence directly, the planner samples a compressed spline representation~\cite{howell_predictive_2022}: candidates are parameterized by knots $\theta$ at fixed time point; the control sequence for each candidate rollout over the horizon is recovered by interpolation, $\mathbf{u}_t = s(t; \theta)$.
As outlined in \cref{alg:cem_planning}, each planner iteration samples $N$ knot vectors from a diagonal Gaussian, evaluates them via parallel rollouts in MuJoCo, and refits the distribution to the $M$ lowest-cost elites. We modify the default CEM planner in MJPC to restrict sampling to the three crane actuators to reduce the search space; the six base degrees of freedom are prescribed from the disturbance prediction module (\cref{subsec:state_estimation}) rather than optimized.

\begin{algorithm}
\caption{Asynchronous Planning with CEM}
\label{alg:cem_planning}
\begin{algorithmic}[1]
\REQUIRE Knot distribution $(\mu, \sigma^2)$, planner iterations $I$, 
sample size $N$, elite count $M$
\WHILE{planning}
    \STATE $\hat{\mathbf{x}}_0 \gets \hat{\mathbf{x}}(t)$ 
    \COMMENT{estimate current state}
    \FOR{$i = 1$ \TO $I$ \COMMENT{planner iterations}}
        \FOR{$j = 1$ \TO $N$ \COMMENT{multi-threaded rollouts}}
            \STATE $\theta^{(j)} \sim \mathcal{N}(\mu, \sigma^2)$ 
            \COMMENT{sample knots (crane actuators only)}
            \STATE $J^{(j)} \gets J(\theta^{(j)}; \hat{\mathbf{x}}_0)$ 
            \COMMENT{evaluate rollout}
        \ENDFOR
        \STATE $\mathcal{E} \gets M$ lowest-cost $\theta^{(j)}$ 
        \COMMENT{rank and select elites}
        \STATE $(\mu, \sigma^2) \gets \text{fit}(\mathcal{E})$ 
        \COMMENT{refit distribution}
        \STATE $\theta^* \gets \arg\min_{\theta \in \mathcal{E}} J(\theta)$ 
        \COMMENT{nominal policy}
    \ENDFOR
    \STATE $\mathbf{u}_t \gets s(t; \theta^*)$ 
    \COMMENT{asynchronous}
\ENDWHILE
\end{algorithmic}
\end{algorithm}

\section{Related Work}
\label{sec:related_work}

\subsection{Classical and Adaptive Anti-Sway Crane Control}

Crane control has been studied extensively, from open-loop filtering to closed-loop feedback strategies~\cite{ramli_control_2017}. Open-loop methods such as input shaping~\cite{vaughan_control_2010} generate reference commands designed to cancel the payload's natural oscillatory modes, and hybrid schemes~\cite{tang_optimization-based_2023, tang_active_2023} combine input shaping with closed-loop MPC to improve robustness. However, these remain limited in offshore settings, where wave motion continuously injects energy into the payload dynamics after the command is shaped.

Closed-loop strategies better address the double-pendulum effect and parametric uncertainties. Non-linear control~\cite{rigatos_nonlinear_2024}, sliding mode control~\cite{li_sliding_2024}, and Lyapunov-based adaptive strategies~\cite{sun_nonlinear_2018, li_adaptive_2024, huang_adaptive_2022} offer formal guarantees on transient performance, ensuring that payload swing and tracking errors converge within predefined bounds despite varying cable lengths. However, these methods are primarily \textbf{reactive}; they suppress sway only after it manifests in the system state.

Conversely, MPC enables \textbf{anticipatory} disturbance rejection by optimizing over a receding horizon. However, applying MPC to coupled, non-linear crane dynamics requires simplifications to keep the online 
optimization tractable, and existing work does so in three ways: \textit{linearization}, approximating the dynamics along a reference path~\cite{bock_real-time_2013} or via Taylor-series expansion to obtain 
convex sub-problems~\cite{rigatos_nonlinear_2024}; \textit{DoF reduction}, omitting axes such as the slew angle~\cite{cao_lyapunov_2023} or decoupling unactuated from actuated dynamics~\cite{lin_constrained_2024}; and \textit{problem decomposition}, separating sway damping from trajectory tracking~\cite{TYSSE2022110219}.
Data-driven variants~\cite{kusznir_nonlinear_2024} rely on identified reduced-order models. All achieve real-time feasibility by discarding coupled slew–luff dynamics, double-pendulum modes or non-linear coupling terms.

\subsection{Learning-Based Approaches}

Recent literature explores deep reinforcement learning (RL) to capture non-linearities without explicit dynamics modeling. Various works apply RL algorithms, such as proximal policy optimization (PPO)~\cite{jang_anti-swing_2023} and soft actor-critic (SAC)~\cite{zhong_learning_2024} for crane control, while Vu et al.~\cite{vu_towards_2025} benchmark RL algorithms for boom-type forestry crane operation. These policies are trained in simulation and, once deployed, generalize poorly to out-of-distribution scenarios~\cite{akki_benchmarking_2025}.
Our MPC-based approach, by contrast, re-optimizes online against the current state rather than executing a fixed policy.

\subsection{Sampling-Based MPC on Robotic Platforms}

The use of high-fidelity simulators for real-time synthesis of complex behaviors has gained significant traction in robotics~\cite{todorov2014drc, alvarez-padilla_real-time_2025, 
pezzato_sampling-based_2025, riviere_optimal_nodate}. 
These approaches leverage parallelized rollouts within physics engines to enable dynamic locomotion~\cite{zhang_whole-body_2025, liang_robust_2025} and manipulation~\cite{li_drop_2025, hess_sampling-based_2025} via sampling-based optimization, and are readily extendable to diverse objects and morphologies without the extensive offline training required by RL~\cite{akki_benchmarking_2025}. Notably, Xue et 
al.~\cite{xue2024fullordersamplingbasedmpctorquelevel} optimize directly over full-order quadruped dynamics in real time without training, avoiding the reduced-order approximations that gradient-based nonlinear MPC typically requires---a motivation shared by our approach. While these applications focus on fully-actuated or contact-rich systems, we extend the paradigm to the underactuated domain of shipboard cranes, where persistent base perturbation couples directly into the unactuated payload dynamics and fast payload transfer conflicts with sway suppression.

\section{Method}
\label{sec:method}

\subsection{Modeling and System Identification}
\label{subsec:sys_id}

Since no MuJoCo model with calibrated actuator dynamics exists for the crane system in \cref{fig:motion_platform}, we construct the model using the physical measurements detailed in \cref{tab:crane_platform_params}. This model serves as the forward dynamics $f(\cdot)$ (see \cref{eq:mpc_objective}) in MJPC. 

The three crane actuators driving the joints $\mathbf{q}_{\text{crane}}$ (slew $\theta$, luff $\gamma$, and hoist $l$ in \cref{fig:system_architecture}) are modeled as velocity actuators in MuJoCo with gain $K_v$ and velocity ranges from \cref{tab:crane_platform_params}, as they are commanded by velocity setpoints. 
Each actuated joint has an armature $I_{arm}$, which represents the rotor inertia of the actuators. 
The cable connecting the boom tip to the payload is modeled as a prismatic slider joint of variable length $l$, assuming that it remains taut. This avoids using a cable or rope in MuJoCo, which increases computation.
For $\mathbf{q}_{\text{payload}}$, we add two pairs of orthogonal hinge joints at the boom tip and payload hook ($x$, $y$ at the boom tip and $\sigma$, $\eta$ at the hook, in \cref{fig:system_architecture}) that simulate the payload's double-pendulum behavior. 
Finally, we attach six-DoF position actuators to $\mathbf{q}_{\text{base}}$ at the crane base, allowing the model to track predicted periodic perturbations during rollouts.

We tune $K_v$ and $I_{arm}$ in two stages: (1)~we adjust $K_v$ to match the commanded steady-state velocity and (2)~we tune $I_{arm}$ to match the transient response slope during acceleration and deceleration. A small damping value is added to slew and luff joint to provide numerical stability during tuning. 
We assess the identified parameters by computing the root-mean-square error (RMSE) between real and MuJoCo joint trajectories over the excitation sequence. The identified parameters and RMSE are reported in \cref{tab:actuator_params}.

\begin{table}[tb]
\centering
\caption{Physical measurements of the crane and motion platform}
\label{tab:crane_platform_params}
\begin{tabular}{lcc}
\toprule
\textbf{Parameter} & \textbf{Value} & \textbf{Unit} \\
\midrule
Boom length & 2.384 & m \\
Slew range & [$-$1.5, 1.5] & rad \\
Luff range & [0, 0.95] & rad \\
Hoist cable range & [0.07, 2.0] & m \\
Slew velocity range & $\pm$0.92 & rad/s \\ 
Luff velocity range & $\pm$0.48 & rad/s \\ 
Hoist velocity range & $\pm$1.0 & m/s \\
Payload mass & 0.317 & Kg \\
Payload dimensions & $\varnothing$0.05$\times$0.46 & m \\
Target diameter & 0.1 & m \\
\midrule
Base $x$, $y$, $z$ displacement & 0.18, 0.0, 0.04 & m \\
Base pitch angle &  [$-$0.16, 0.13] & rad \\
Period (\textsc{slow} / \textsc{medium} / \textsc{fast}) & 12 / 7 / 5 & s \\
\bottomrule
\end{tabular}
\end{table}

\begin{table}[]
    \centering
    \caption{Identified actuator and joint parameters, with RMSE against 
    measured crane trajectories.}
    \label{tab:actuator_params}
    \begin{tabular}{l|ccc|c}
    \toprule
    \textbf{Joint} & \textbf{$K_v$} & \textbf{$I_{arm}$} 
    & \textbf{Damping} & \textbf{RMSE} \\
    \midrule
    Slew  & 7800  & 1000 & 0.01 & $1.5\times10^{-3}$\,rad \\
    Luff  & 13000 & 2200 & 0.01 & $9.5\times10^{-5}$\,rad \\
    Hoist$^\dagger$ & 25000 & 3200 & 0.0 & $1.1\times10^{-2}$\,m \\
    \bottomrule
    \multicolumn{5}{l}{\footnotesize $^\dagger$Add \texttt{frictionloss} = 30 to prevent 
    payload slip} \\
    \end{tabular}
\end{table}

\subsection{Cross-Entropy Method for Model Predictive Control}
\label{subsec:cem}

\subsubsection{Online Planning for Real-Time Control}
\label{subsubsec:cem_realtime}

As outlined in \cref{alg:cem_planning}, in each planning cycle, the planner runs $I=5$ iterations to refine the policy before the first action of the nominal trajectory is executed on the 
crane. The distribution warm-starts across both the $I$ iterations and successive planning cycles. Because the state estimate is held fixed across the $I$ iterations, planning proceeds asynchronously from execution: the planner refines the policy as fast as compute allows, at a rate higher than the crane control frequency of 20\,Hz. The prediction horizon is 0.8\,s; extending it further does not achieve measurable improvement in sway damping while significantly reducing the frequency of planning. \cref{tab:cem_params} summarizes the CEM hyperparameters.

\begin{table}[tb]
\centering
\caption{CEM hyperparameter values}
\label{tab:cem_params}
\begin{tabular}{cc}
\toprule
\textbf{Parameter} & \textbf{Value} \\
\midrule
Horizon length ($H$) & 80 steps (0.8 s) \\
Model time-step ($\Delta t$) & 0.01 s \\
Planner iterations ($I$) & 5\\
Trajectory sample size ($N$) & 20 \\
Elite count ($M$) & 5 \\
Sampling noise ($\sigma$) & 0.2 \\
No. of spline knots ($K$) & 3 \\
Spline representation & Zero-order hold \\ 
\bottomrule
\end{tabular}
\end{table}

\subsubsection{Cost Function Design}
\label{subsubsec:cost_func}

The cost function in \cref{eq:mpc_objective} must reconcile two competing objectives~\cite{sun_nonlinear_2018}: accurately tracking a moving target and maintaining minimal payload oscillation over the target position. 
The individual cost terms are detailed in \cref{tab:cost_terms}. 

To ensure smooth station-keeping, $r_{\text{target}}$ employs a Pseudo-Huber norm. Its flat region near zero suppresses actuator jitter once the payload is within an acceptable tolerance of the goal, where positioning is de-prioritized in favor of active sway damping.
Oscillation cost terms sway damping $r_{\text{sway}}$ and payload tilt $r_{\text{tilt}}$ grow linearly to prevent disproportionate or destabilizing corrective effort during large transient swings.
The quadratic norm applied to $r_{\text{vel}}$ and $r_{\text{ctrl}}$ penalizes unsafe magnitudes super-linearly.

During tuning with the crane system, no single fixed weighting satisfied both objectives: emphasizing $r_{\text{target}}$ and $r_{\text{sway}}$ achieved accurate station-keeping but overshot on a moving target, while emphasizing $r_{\text{vel}}$ prevented overshoot at the cost of sustained oscillation near the goal. 
We resolve this with a distance-dependent $\tanh$ blend that adjusts the weights $\alpha$ and $\beta$ in \cref{tab:cost_terms}. It interpolates between the two regimes as a continuous relaxation of discrete cost-function switching~\cite{todorov2014drc}.
Let 
$d = \lVert\mathbf{p}_{\mathrm{target}} - \mathbf{p}_{\mathrm{payload}}\rVert$ be the distance to the target:
\begin{align}
  \alpha(d) &= \tfrac{1}{2}\bigl(\tanh\!\bigl(k_d\,
  (d - d_{\mathrm{th}})\bigr) + 1\bigr), 
  \label{eq:alpha}\\
  \beta(d)  &= \tfrac{1}{2}\bigl(\tanh\!\bigl(-k_e\,
  (d - d_{\mathrm{th}})\bigr) + 1\bigr) + 1, 
  \label{eq:beta}
\end{align}
where $d_{\mathrm{th}} = 0.1$\,m is the proximity threshold and $k_d = 10$, $k_e = 5$ control transition sharpness. Far from the target ($\alpha \!\to\! 1$, $\beta \!\to\! 1$), $r_{\text{target}}$ and 
$r_{\text{sway}}$ dominate to guide the payload while suppressing oscillation. Within the proximity region ($\alpha \!\to\! 0$, $\beta \!\to\! 2$), emphasis shifts to velocity matching, decelerating the payload to prevent overshoot.

$r_{\text{ctrl}}$ penalizes normalized slew and luff velocities with unit weight. 
Its weighted contribution is far below that of the blended $r_{\text{sway}}$ and  $r_{\text{vel}}$ terms. In our experiments, setting its weight to zero produced no measurable change in either the actuator commands or the sway-damping performance on the moving platform. We retain it as a weak regularizer and omit hoist, whose inclusion also has negligible effect on the policy.

We add a terminal cost $c_f$ in \cref{eq:mpc_objective} which is an unweighted penalty on the payload's terminal distance to the target.

\begin{table}[tb]
\centering
\caption{Cost function terms}
\label{tab:cost_terms}
\begin{tabular}{llc}
\toprule
\textbf{Term} & \textbf{Formulation} & \textbf{Weight}\\
\midrule
$r_{\text{target}}$: target tracking  & $\sqrt{\|\mathbf{p}_{\text{payload}} - \mathbf{p}_{\text{target}}\|^2 + \epsilon^2} - \epsilon$ & $A\alpha(d)$ \\
& \footnotesize with $\epsilon=0.05$ & \\
\addlinespace
$r_{\text{sway}}$: sway damping  & $\sqrt{|\theta_{\text{sway}}|^2 + \delta^2}$ & $\frac{A}{2}\alpha(d)$ \\
& \footnotesize with $\delta=2.0$ & \\
\addlinespace
$r_{\text{vel}}$: relative velocity  & $\|\dot{\mathbf{p}}_{\text{payload}} - \dot{\mathbf{p}}_{\text{platform}}\|^2$ & $B\beta(d) $ \\
\addlinespace
\midrule
$r_{\text{ctrl}}$: control effort  & $\|\bar{\mathbf{u}}_{\mathrm{slew}}\|^2, \|\bar{\mathbf{u}}_{\mathrm{luff}}\|^2$ & 1\\
\addlinespace
$r_{\text{tilt}}$: payload tilt  & $\sqrt{|2\arccos(|\mathbf{q}_{\text{payload}} \cdot \mathbf{q}_{\uparrow}|)|^2 + \delta^2}$ & 500\\
& \footnotesize with $\delta=3.0$ & \\
\bottomrule
\end{tabular}
\\[4pt]
\footnotesize Note: $\theta_{\text{sway}} = \arcsin(\frac{\sqrt{d_x^2+d_y^2}}{\lVert\mathbf{b}\rVert})$ where $\mathbf{b}$ is the boom-tip-to-payload vector. $A=200, B=350$. \\
\end{table}

\subsection{State Estimation and Disturbance Prediction}
\label{subsec:state_estimation}

Our state estimation consists of two parts: instantaneous estimation of the crane-payload system state and prediction of the moving platform’s pose.
We obtain the joint positions $\mathbf{q}_{\text{crane}}$ from actuator encoders at 20\,Hz. We measure payload angles $\mathbf{q}_{\text{payload}}$ with a motion capture system at 100\,Hz, which provides reliable ground-truth pose estimation. 
We compute all the joint velocities by applying a $10$-step moving average filter to the finite-differenced joint positions in \eqref{eq:system_state}.

We inject the predicted $\mathbf{q}_{\text{base}}$ sequence over the entire horizon length $H$ directly into the MuJoCo forward rollouts, bypassing the need for analytical disturbance modeling. Because the planner requires only a forecast of the base pose over the prediction horizon, it can anticipate how the base excitation will perturb the payload during rollout and plan damping actions accordingly.

We treat the base perturbations $\mathbf{q}_{\text{base}}$ introduced by the motion platform as periodic trajectories, forecasting them via autocorrelation-based pattern matching over a two-period sliding window of motion-capture data of $\mathbf{q}_{\text{base}}$. Residual prediction errors are corrected at each re-planning cycle by updating the instantaneous state and disturbance estimates. Crucially, this periodicity assumption is confined to the prediction module: since the planner accepts any horizon forecast, a non-periodic or stochastic predictor could supply the same sequence without altering the control pipeline. Our current experiments use only this periodic predictor, and validating the pipeline with non-periodic disturbance forecasts is an important next step.

\section{Experimental Setup}
\label{sec:experimental-setup}

\subsection{Hardware Setup}
\label{subsed:hardware-setup}

We perform all experiments on the setup in \cref{fig:motion_platform}.
The crane receives slew, luff and hoist velocity commands and publishes the actuator encoder states at 20\,Hz using ROS. 
To demonstrate the feasibility of our MJPC controller across different compute regimes, we run it on two platforms: (1) a high-performance laptop (Intel Core i9-12900HK CPU) and (2) an edge computing module (NVIDIA Jetson AGX Orin 64GB). Both platforms use the same CEM parameters in \cref{tab:cem_params} and differ only in planning frequency (${\sim}50$\,Hz on the laptop, ${\sim}40$\,Hz on the Jetson).

\subsection{Baselines}

We evaluate our controller (MJPC) against a classical baseline (PID) and a learning-based baseline (RL).

\textbf{Proportional-integral-derivative (PID).} We implement a PID controller for simultaneous target position tracking and payload sway damping. The controller combines a PD loop for joint position tracking with a nested PID loop dedicated to payload sway damping for slew, luff and hoist actuator. This represents a reactive classical baseline that lacks the predictive horizon of our approach.

\textbf{Proximal policy optimization (PPO).} While both PPO and SAC have been applied to crane control~\cite{jang_anti-swing_2023, zhong_learning_2024}, we adopt PPO~\cite{schulman2017proximal} as a representative model-free baseline, following recent MPC-versus-RL benchmarks that favor PPO for locomotion control task~\cite{akki_benchmarking_2025}.
To isolate the comparison between control architectures rather than robustification pipelines, both MJPC and PPO operate from the same system-identified model~(\cref{subsec:sys_id}) without domain randomization.
Domain randomization could improve the robustness of either approach---the learned policy through randomized training, and the planner through ensemble or randomized rollouts---but adding it to either side would confound the architectural comparison. We therefore withhold it from both.

We train the agent with a payload disturbance---a force applied at the start of each episode to excite the swing dynamics---but not with the periodic base motion used at test time, since MJPC has no offline training phase and predicts the base motion online during planning.
Both controllers instead first encounter the base excitation at deployment---MJPC through its horizon forecast, PPO through its state history---reflecting the intended contrast between anticipatory and reactive control.
The policy network uses three hidden layers of $256$ neurons.

In preliminary experiments, we tested two reward designs: one following the cost function from \cref{subsubsec:cost_func}, as well as a handcrafted reward penalizing payload distance from the target and acceleration.
Across $10$ seeds each, we observed that all policies converged to ``bang-bang'' control, outputting actions of maximum $-1$ and $+1$ velocity. 
While this strategy performed well in simulation, deployment on the real crane proved unstable, demonstrating progressively increasing oscillations.
In our final experiments, we scale actions by $0.25$ to mitigate this effect and deploy the best performing policy, using our handcrafted reward.

We report statistical significance for numerical values using Mann-Whitney U tests~\cite{mann1947test}.

\section{Results}
\label{sec:results}

We evaluate three aspects of the MJPC controller: its sway-damping and tracking performance under base perturbation (\cref{subsec:sway_damping_main}), the real-time feasibility of CEM planning under compute constraints (\cref{subsec:cem_ablations}), and its robustness to model discrepancy (\cref{subsec:robustness_tests}).

\begin{table*}[tb]
    \centering
    \caption{
    Summary statistics of mean position error and payload tilt over the 10--20\,s station-keeping interval, over 10 replications. In the top four rows, bold marks results significantly better than both baselines; two bold values are jointly best, not significantly different from each other. The bottom three rows ($\Delta$) report each controller's error increase from its single-payload baseline under an unmodeled double payload, over four replications.
    }
    
    \begin{tabular}{lcccccccc}
        \toprule
        Method & \multicolumn{2}{c}{Static} & \multicolumn{2}{c}{Slow} & \multicolumn{2}{c}{Medium} & \multicolumn{2}{c}{Fast} \\
        \cmidrule(lr){2-3} \cmidrule(lr){4-5} \cmidrule(lr){6-7} \cmidrule(lr){8-9}
         & Pos (m) & Ang (deg) & Pos (m) & Ang (deg) & Pos (m) & Ang (deg) & Pos (m) & Ang (deg) \\
        \midrule
        RL & 0.27 (0.40) & 7.85 (3.89) 
        & 0.39 (0.19) & 9.92 (2.33) 
        & 0.47 (0.12) & 13.82 (2.80) 
        & 0.43 (0.12) & 13.25 (2.01) \\
        PID & 0.08 (0.03) & 1.42 (0.11) 
        & 0.22 (0.07) & 1.75 (0.78) 
        & 0.16 (0.04) & 3.66 (0.87) 
        & 0.20 (0.18) & 3.61 (26.65) \\
        MJPC (ours) & \textbf{0.03 (0.01)} & 1.86 (1.32) 
        & 0.09 (0.01) & 1.51 (0.34) 
        & 0.11 (0.02) & \textbf{1.75 (0.78)} 
        & \textbf{0.11 (0.02)} & \textbf{2.21 (1.24)} \\
        MJPC [edge] (ours) & \textbf{0.03 (0.01)} & 1.44 (0.77) 
        & \textbf{0.06 (0.02)} & 2.15 (0.75) 
        & \textbf{0.07 (0.02)} &\textbf{ 2.57 (0.38)} 
        & \textbf{0.10 (0.02)} & 3.47 (1.02) \\
        \midrule
        RL $\Delta$ & 0.42 & 3.04 
        & 0.35 & 3.86 
        & -0.04 & 0.27 
        & 0.48 & 2.53 \\
        PID $\Delta$ & 0.11 & 0.22 
        & 0.06 & 0.78 
        & 0.15 & -1.39
        & 0.13 & -0.52 \\
        MJPC $\Delta$ [edge] (ours) & 0.03 & 2.06 
        & 0.17 & 2.79 
        & 0.23 & -0.76
        & 0.16 & 0.28 \\
        \bottomrule
    \end{tabular}
    \label{tab:ab_target_error}
\end{table*}

\subsection{Payload Sway Damping and Target Tracking under Perturbation}
\label{subsec:sway_damping_main}

\begin{figure}[tb]
     \centering
     \includegraphics[width=\linewidth]{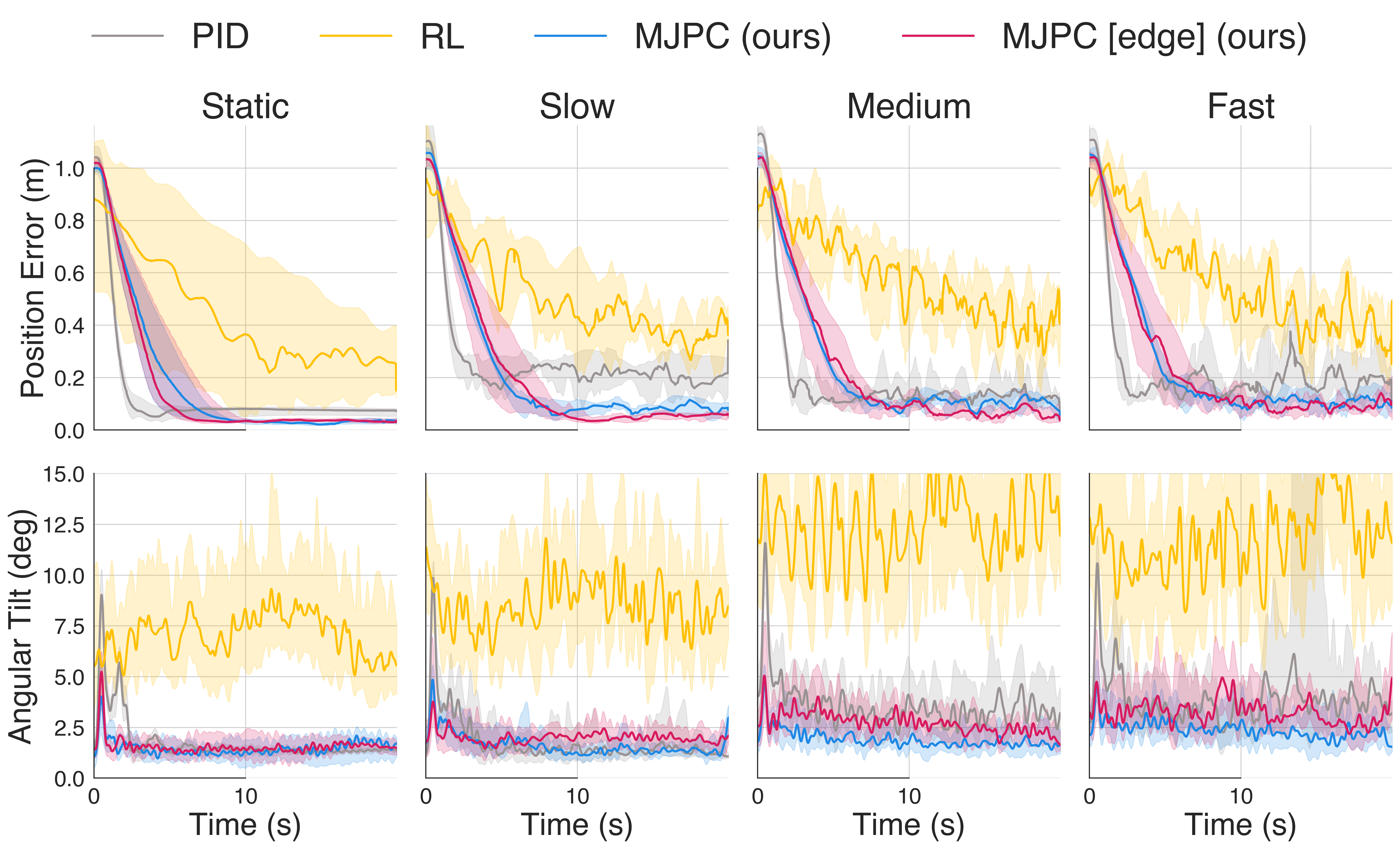}
     \caption{Payload position error from target (top row) and angular tilt (bottom row) over various motion platform settings. We plot the median (solid line) and IQR (shaded area) over 10 replications.}
     \label{fig:main_results_traj}
\end{figure}

To evaluate the performance of our proposed MJPC controller, we design a target switching task using two targets mounted $1$\,m apart on the motion platform, shown in \cref{fig:motion_platform}. Starting from \textsc{Target 1}, we allow $20$\,seconds for the controller to move the payload to \textsc{Target 2} while damping sway, after which we swap targets and repeat. We collect data from $10$ trajectories between the two targets with a total duration of $200$\,seconds on \textsc{static}, \textsc{slow}, \textsc{medium} and \textsc{fast} motion platform settings. 

In \cref{fig:main_results_traj}, we display aggregated trajectories over all replications of the payload's $xy$ position error relative to the target (top) and the tilt angle between the payload longitudinal axis and the global $z$-axis (bottom).
Since the payload is given 20\,s to travel between targets, we discard the first 10\,s of approach and aggregate the mean $xy$ position error over the $10$--$20$\,s interval 
(\cref{tab:ab_target_error}), measuring each controller's station-keeping ability over the target.

Our evaluations show that MJPC successfully reconciles the conflicting objectives of dynamic target tracking and payload sway suppression. 
The RL baseline performs poorly on hardware despite training on the system-identified model. During training, the PPO agent converges to outputting maximum velocity actions as a local optimum to the task reward function. This highlights a common limitation of model-free control when the learned policy cannot adapt at deployment. Future work could explore tuning the reward function to mitigate the observed ``bang-bang'' control effect. 

The classical PID reactive baseline degrades substantially as the external disturbance induced by the motion platform increases, failing to anticipate continuous base perturbations. While PID often drives the payload to reach the target position faster, it does so at the cost of inducing more severe oscillations on the payload over the target.

Conversely, MJPC demonstrates the best overall performance with least degradation across different platform speeds. In particular, despite its lower planning frequency, the Jetson performs comparably to the laptop variant. Because both exceed the 20\,Hz crane control frequency, each supplies a freshly refined plan before every control step, so the additional planning headroom on the laptop is not consumed by the control loop. The two platforms therefore run an effectively identical controller, and any residual differences in \cref{tab:ab_target_error} are not attributable to planning quality, which uses identical parameters across platforms.

\subsection{Feasibility of Real-Time CEM Planning}
\label{subsec:cem_ablations}

\begin{figure}[tb]
     \centering
     \includegraphics[width=\linewidth]{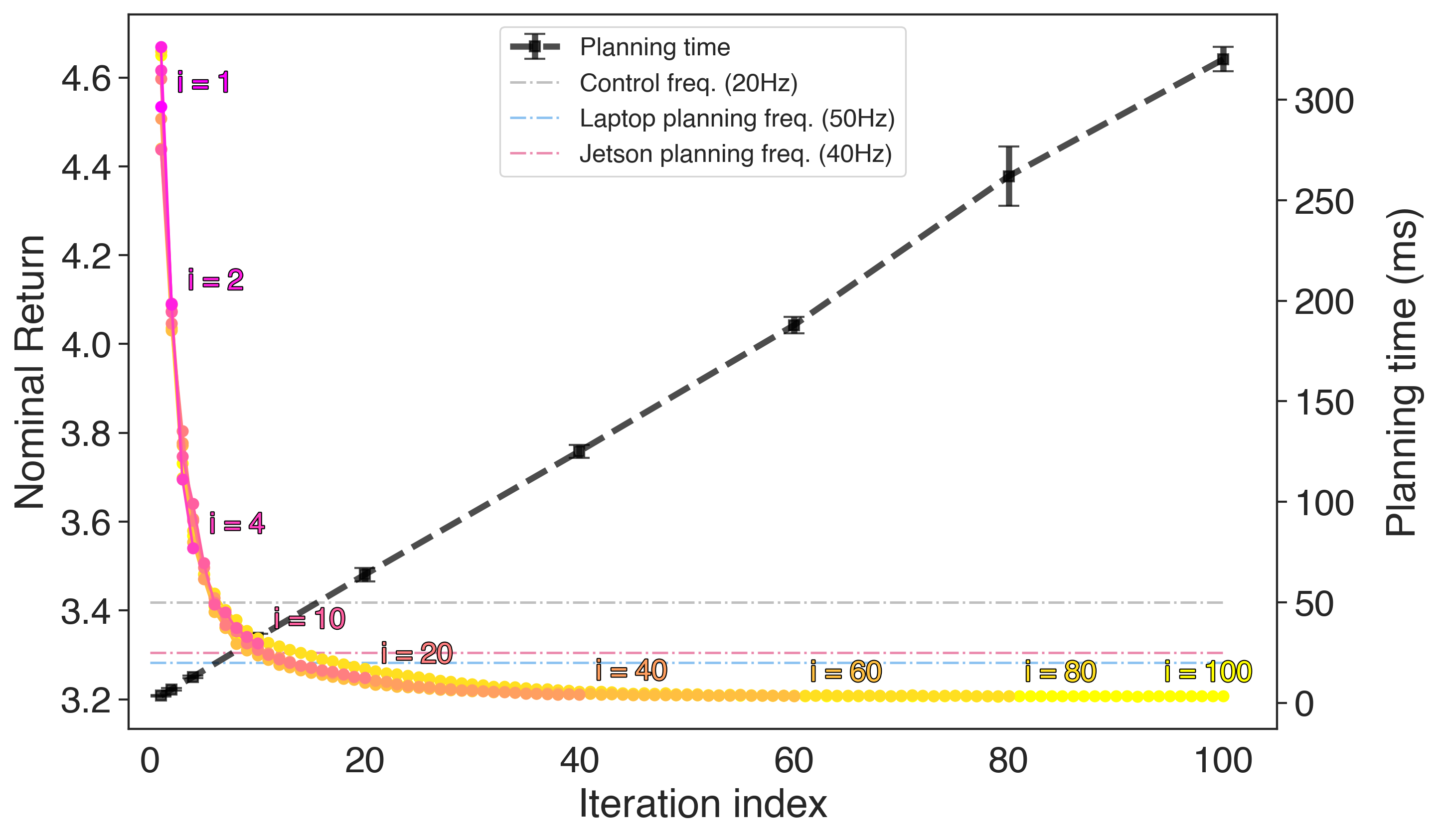}
     \caption{Mean nominal return over 50 replications for various CEM planner iterations $I$ in simulation. We also plot the planning time per cycle for each iteration parameter.}
     \label{fig:planner_step_sim}
\end{figure}

We first analyze convergence of the CEM planner entirely in simulation, where the planner is commanded to run a number of planner iterations ($I$) (ranging from $1$ to $100$) to optimize the policy for the next step. For each $I$, we average the nominal return (lowest trajectory cost prior to the action update) over $50$ independent runs. 
As shown in \cref{fig:planner_step_sim}, the nominal return decreases sharply and plateaus after approximately $60$ iterations, confirming convergence. 
Planning time, however, increases linearly with each iteration (dashed line): beyond $10$ iterations the planner cannot complete a cycle within the 20\,Hz control period, so subsequently on real hardware it would lag the current crane state.

\begin{figure}[tb]
     \centering
     \includegraphics[width=\linewidth]{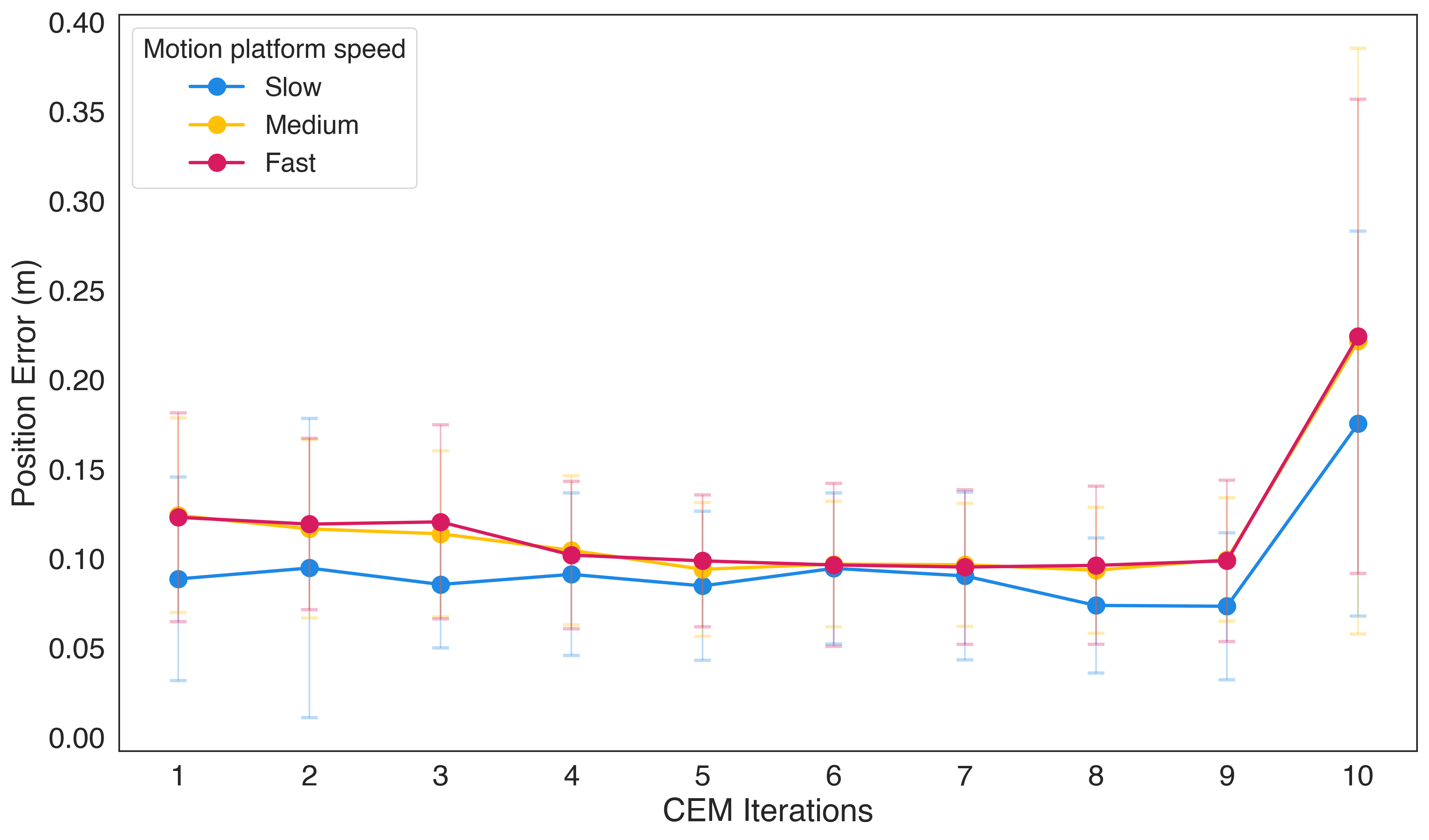}
     \caption{Mean payload position error from target, on the physical crane, over 60\,s interval for various CEM planning iterations. The range is capped at 10 iterations, beyond which planning time exceeds the 20\,Hz minimum control period.}
     \label{fig:planner_step_real_crane}
\end{figure}

To establish practical feasibility, we replicate the above on the physical crane, but from $1$ to $10$ $I$ only. For each $I$, we deploy the controller to maintain the payload over \textsc{Target 1} and record the $xy$ position error over $1$ minute at different platform speeds. As shown in the result in \cref{fig:planner_step_real_crane}, too few planner iterations produce under-converged policies with high tracking error, while too many introduce latency that destabilizes the system. Operating at $I=5$ strikes the balance: the resulting replanning rate (${\sim}40$\,Hz) provides sufficient policy refinement to suppress payload sway while remaining well above the 20\,Hz crane control frequency (see the real-time bounds in \cref{fig:planner_step_sim}).

\subsection{Robustness under Model Discrepancies}
\label{subsec:robustness_tests}

Finally, we assess MJPC's robustness to mismatch between the MuJoCo planning model and the physical system, a scenario common in crane operations where load mass distribution shifts during transfer. Replicating the target-tracking setup of \cref{subsec:sway_damping_main}, we attach a secondary cylindrical payload (0.35\,m, 0.23\,kg) beneath the primary so both swing freely, forming a triple-pendulum system. The controller retains the unmodified double-pendulum model, leaving the 74\% added mass and the extra DoF entirely unmodeled. Owing to increased payload instability, we collect four replications per controller per setting. This introduces two disturbance modes absent from the planning model: out-of-phase oscillation between the payloads during platform movement, and, at higher speeds, circular swinging that induces rotational spin on the primary 
payload.

\cref{tab:ab_target_error} reports the error increase from the single-payload setting. 
PID shows the smallest angular degradation, its sway \emph{improving} at \textsc{medium} and \textsc{fast} platform speeds ($-1.39$\textdegree, $-0.52$\textdegree): its low-bandwidth actuation avoids exciting the secondary 
pendulum mode, letting the extra payload act as passive damping. This comes at the cost of increased station-keeping error (0.06--0.15\,m above nominal), not seen in the single-payload case. RL degrades most, with position error rising 0.35--0.48\,m and angular error up to $3.86$\textdegree, as the learned policy, trained on nominal payload dynamics, fails to adapt to the added complexity in the system.

On the other hand, MJPC retains the lowest position error across all settings despite a 0.03--0.23\,m rise, degrading gracefully as frequent replanning corrects positional error through state feedback. Its angular error increases most at \textsc{static} and \textsc{slow} speeds ($+2.06$\textdegree, $+2.79$\textdegree), reflecting persistent out-of-phase oscillation that replanning cannot fully suppress; at \textsc{fast} platform speed, circular swinging occasionally developed into sustained spin that interrupted individual trials.

\section{Conclusion and Future Work}

We present a real-time, sampling-based MPC framework for shipboard boom crane control that pairs a system-identified MuJoCo model with a CEM planner to suppress double-pendulum sway under base perturbations. By evaluating candidate policies through rollouts in simulation, our approach sidesteps the linearization, DoF reduction, or analytical model derivation common in existing crane MPC, while an adaptive cost function reconciles fast payload transfer with sway suppression. Hardware experiments confirm that this approach performs effectively under strict planning-time constraints, making simulation-based planning feasible on compute-constrained hardware.

Our evaluation compares against classical (PID) and learned (RL) baselines but not against linearized or reduced-order MPC. While our approach retains coupled nonlinear dynamics that such methods discard, a direct comparison against linearized or reduced-order MPC baselines, which would quantify the benefit of retaining coupled nonlinear dynamics, remains an important next step.
We plan to replace the periodic disturbance assumption with a predictive state estimator such as a Kalman filter, and to evaluate the controller under injected prediction errors such as phase drift and amplitude mismatch.
For deployment in unstructured outdoor conditions, we will also eliminate reliance on motion capture through onboard sensor fusion (IMUs, RGB-D or LiDAR), a transition that introduces estimation noise and inference latency, likely requiring planner retuning to absorb. 
Finally, we aim to extend the MuJoCo model with an additional hinge joint capturing the payload's rotational spin about the cable axis, currently unmodeled in the double-pendulum representation.






\section*{ACKNOWLEDGMENT}
We want to thank the members of our performer team at the LINC program, namely the members of Peraton Labs and the Safe Robotics Laboratory from Princeton University, for the fruitful discussions and their feedback.


\bibliographystyle{IEEEtran}
\bibliography{IEEEtranBST/IEEEabrv, bibliography}

\end{document}